\documentclass{article}
\usepackage{spconf,amsmath,graphicx}
\usepackage{bm}
\usepackage{bbm}
\usepackage{amssymb}
\usepackage{booktabs}
\usepackage{xcolor}
\usepackage{soul}
\newcommand{\vet}[1]{\bm{\mathrm{#1}}}

\newcommand{\etal}{\textit{et al}. }
\newcommand{\ie}{\textit{i}.\textit{e}., }
\newcommand{\eg}{\textit{e}.\textit{g}., }

\newcommand{\rul}[2][white]{\setulcolor{#1}\ul{#2}\setulcolor{black}}
\title{Unsupervised \rul{Domain-Adaptive} Person Re-identification \\Based on Attributes}
\name{Xiangping Zhu, Pietro Morerio and Vittorio Murino\thanks{We gratefully acknowledge the support of NVIDIA Corporation with the donation of the Titan Xp GPU used for this research.}}
\address{Pattern Analysis and Computer Vision \rul{(PAVIS)}\\ Istituto Italiano di Tecnologia, Genova, Italy}
\begin{document}
%
\maketitle
\begin{abstract}
Pedestrian attributes, \eg hair length, clothes type and color, locally describe the \rul{semantic appearance} of a person. Training person re-identification (ReID) algorithms under the supervision of such attributes have proven to be effective in extracting local features which are important for ReID. Unlike person identity, attributes are consistent across different domains (or datasets). However, most of ReID datasets lack attribute annotations. On the other hand, there are several datasets labeled with sufficient attributes for the case of pedestrian attribute recognition. Exploiting such data for ReID purpose can be a way to alleviate the shortage of attribute annotations \rul{in ReID case}. In this work, an unsupervised domain adaptive ReID feature learning framework is proposed to make full use of attribute annotations. We propose to transfer attribute-related features from their original domain to the ReID one: \rul{to this end, we introduce an adversarial discriminative domain adaptation method in order to learn domain invariant features for encoding semantic attributes}. \rul{Experiments on three large-scale datasets validate the effectiveness of the proposed ReID framework.}
\end{abstract}

\begin{keywords}
Person re-identification, unsupervised domain adaptation, unsupervised person re-identification, pedestrian attributes
\end{keywords}

\section{Introduction}
\label{sec:intro}

Person ReID aims at matching a target person across cameras. This is still a challenging problem due to \rul{large intra-personal variations and inter-personal similarities} \cite{zheng2016reid_survey, GMIE}. For these reasons it is of paramount importance to have a model that generates robust and consistent features modeling all these kind of variations. \rul{Pedestrian attributes are semantically meaningful for describing person appearance. In addition, attributes are also consistent between different domains. Based on these observations, as in Fig. 1, we propose an unsupervised domain-adaptive framework which learns attribute-related features from an annotated source dataset and adapts them for ReID in a target dataset}.

Different from person identity, which denotes the global description of the person, \rul{attributes represent local parts of a person}. For example, hair length refers to the head part, while the \rul{upper-body} clothing type mainly focuses on the torso. Many works have demonstrated that \rul{ReID algorithms} can benefit a lot from person attribute supervision \cite{layne2012attribute, lin2017attribute_recognition, matsukawa2016person}, however, \rul{most ReID datasets} come without person attributes \cite{zheng2016reid_survey, cuhk03, zheng2016mars, wei2017msmt17}. In addition, compared with producing identity labels, attribute labeling is a much more complex and time consuming task. Despite the effort of Lin \etal \cite{lin2017attribute_recognition} in annotating few ReID datasets with \rul{attribute} labels, there is still a big shortage of attribute-annotated large-scale ReID datasets. On the other hand, there is a number of attribute-labeled datasets in the domain of pedestrian attribute recognition \cite{RAP, attribute_dataset1, li2016wider_attribute_dataset}. Unfortunately, this kind of datasets cannot be directly exploited for ReID, since no identity label is provided and usually there is only one image for each person.

\begin{figure}[t]
\centering
\includegraphics[width=0.40\textwidth]{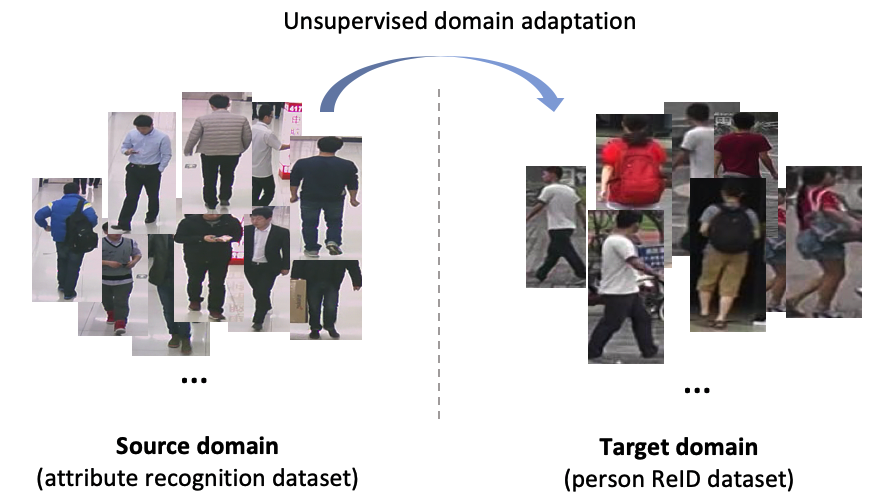}\\
\caption{Overview of the proposed unsupervised domain adaptive ReID framework. \rul{The model trained on source domain is adapted to the target domain. Source domain is labeled with attributes, while in the target one, label is not needed.} The images are picked from RAP (attribute recognition dataset) \cite{RAP} and Market1501 (ReID dataset) \cite{market1501}.}
\label{fig: proposed_reid_framework}
\end{figure}

\begin{figure}[t]
\centering
\includegraphics[width=0.48\textwidth]{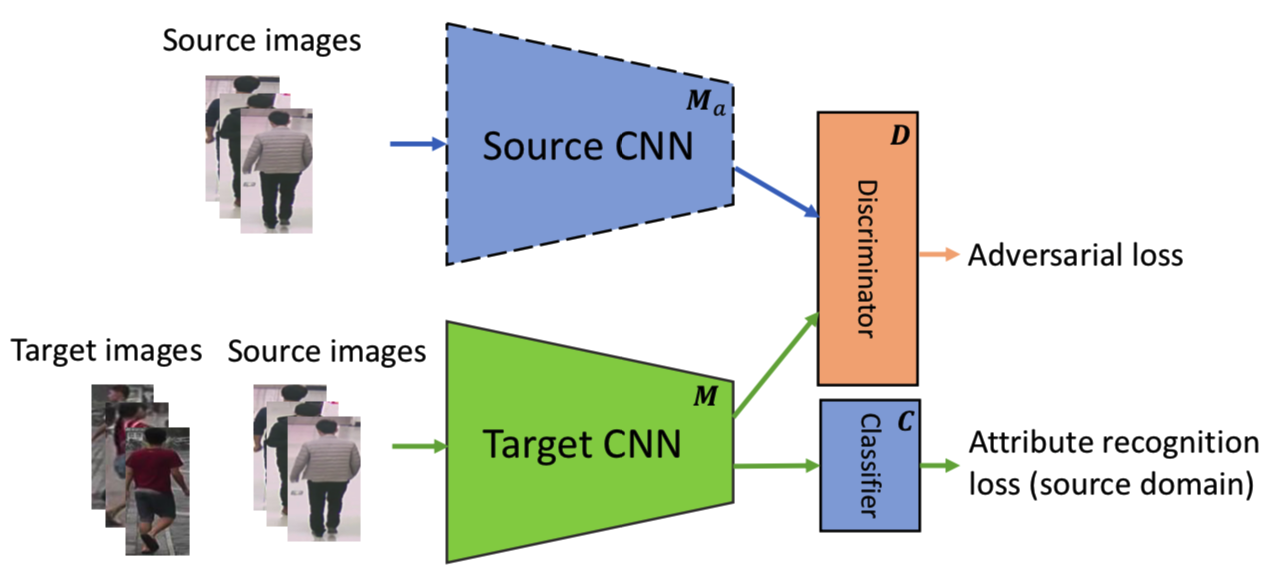}\\
\caption{The proposed unsupervised adversarial adaptation. The source CNN is pretrained on attribute recognition dataset. The classifier is only used for attribute \rul{recognition} on the source images. \rul{Dashed lines indicate fixed network weights.}}
\label{fig: adversarial adaptation}
\end{figure}

As shown in Fig. \ref{fig: proposed_reid_framework}, \rul{our work bridges the gap between attribute recognition and person ReID. In the proposed unsupervised domain-adaptive ReID framework, an attribute recognition model, actually, a convolutional neural network (CNN), is first trained on attribute recognition dataset (denoted as the source domain). Considering no attribute labels are available in ReID dataset (target domain), an unsupervised adversarial domain adaptation method, based on the trained attribute recognition model, is then applied for learning domain invariant attribute-related features.} Fig. \ref{fig: adversarial adaptation} \rul{details the proposed domain adaptation method which is extended from the adversarial discriminative domain adaptation (ADDA)} \cite{tzeng2017adda, volpi2017adversarial}. \rul{Specifically, as in Fig. 2, both the source and target samples are fed into the target CNN to make it invariant to these two domains. In addition, in order to maintain the attribute recognition performance during the adversarial adaptation, an additional attribute classifier is added along with the discriminator. After adaptation, the attribute-related features in ReID domain are extracted using the target CNN. This kind of features proved to be very effective such that the simple Euclidean distance is sufficient to estimate the similarity for matching or ranking person images, unlike more complex metric learning approaches} \cite{zheng2016reid_survey,zheng2016mars}. \rul{To evaluate the effectiveness of the proposed  ReID framework, three datasets have been tested, namely the attribute recognition dataset RAP} \cite{RAP} and two ReID datasets Market-1501 \cite{market1501} and DukeMTMC-reID \cite{DukeMTMC-reID}.

\section{Related works}
\label{sec:related_works}

This work mainly focuses on attribute based person ReID and unsupervised domain adaptation. The effectiveness of using attributes for ReID has been reported in  many works \cite{zheng2016reid_survey, layne2012attribute, lin2017attribute_recognition, matsukawa2016person, su2016deep}. Lin \etal jointly consider the ID and attribute as the supervisions in a deep multi-task model, to learn the discriminative ReID features \cite{lin2017attribute_recognition}. Similarly, \cite{wang2018transferable, lin2018multi} also jointly consider person ID and attributes for ReID but under the unsupervised domain adaptive framework. Matsukawa \etal only consider the attribute as the supervision to learn the feature representation, and, in addition, a combination loss is designed to improve the feature discrimination \cite{matsukawa2016person}. In \cite{su2016deep}, a deep attribute learning algorithm has been introduced, and the attribute recognition knowledge is transferred from one dataset to another one using a fine tuning strategy.

Unsupervised domain adaptation is proposed to solve the domain shift problem when label information is not provided in some target domain. The basic idea behind lots of domain adaptation works is to match the feature distributions in the source and target domains \cite{uda_backpropagation, domain_adaptation_survey, volpi2017adversarial}, or their statistics \cite{morerio2018minimalentropy}.  Ganin \etal proposed a unsupervised domain adaptation method that can simultaneously train the network to learn discriminative and domain invariant feature representations by using a gradient reversal layer \cite{uda_backpropagation}. In recent years, with the emergence of generative adversarial networks (GANs) \cite{gans}, there are also some works trying to using adversarial adaptation strategy for adapting the network from source to target domain. Tzeng \etal proposed an adversarial adaptation framework in which the network is \rul{first} trained in the source domain and then adversarially \rul{adapted} to the target domain with the help of a discriminator as in GANs \cite{gans, tzeng2017adda}. The work \cite{volpi2017adversarial} extends this adversarial adaptation framework with adding the step to train a feature generator.

\section{\rul{Methodology}}
\label{sec:methodology}

Suppose there are $n_a$ samples from the attribute recognition (source) domain $\mathcal{D}_a = \{(\vet{x}^a_i, \vet{a}_i)\}_{i=1}^{n_a}$, in which $\vet{x}_i^a$ represents the $i$th sample, featuring $m$ attributes $\vet{a}_i \in \mathbb{R}^{m}$. For the ReID (target) domain $\mathcal{D}_p = \{\vet{x}_i^p \}_{i=1}^{n_p}$, there are $n_p$ samples $\vet{x}^p_i$, but, on the contrary, no annotation is available. We want to learn a domain invariant mapping $\vet{M}$, which can be used to extract attribute-related features for ReID.

In the proposed framework, depicted in Fig. \ref{fig: adversarial adaptation}, CNN is used to learn the mapping $\vet{M}$. The network is \rul{first} trained in the attribute recognition domain $\mathcal{D}_a$ and then adapted to the ReID domain $\mathcal{D}_p$. An unsupervised domain adaptation method, based on \cite{tzeng2017adda}, is used. As already mentioned, no attribute label is available in the ReID domain. To learn the domain invariant feature representation, both the source and target images are fed into a target CNN as shown in Fig. \ref{fig: adversarial adaptation}, which is adversarially trained with a discriminator $\vet{D}$. In addition, in order to ensure the attribute recognition performance during the domain adaptation, a classifier is added on the target CNN (for source images only).

\begin{figure}[t]
\centering
\includegraphics[width=0.40\textwidth]{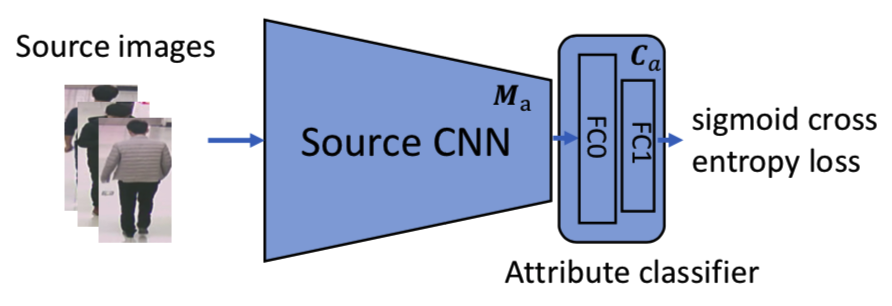}\\
\caption{The attribute recognition network architecture.}
\label{fig: attribute_recognition}
\end{figure}

\vspace{2mm}
\noindent
\textbf{Attribute Recognition:} As shown in Fig. \ref{fig: attribute_recognition}, the attribute recognition is treated as a multi-label classification task. Given one sample $\vet{x}_i^a \in \mathcal{D}_a$, there are $m$ attributes $\vet{a}_{i}=\{a_{i,1}, \ldots, a_{i, m}\} \in \mathbb{R}^{m}$. Thus, instead of using softmax cross entropy loss, the sigmoid cross entropy loss is used to train the network:
\begin{equation}
\label{eq:ar_loss}
	\begin{split}
		L_{attr} = & - \mathbb{E}_{\vet{x}_i \sim \vet{X}^{a}} \sum_{j=1}^{m} \Big(a_{i,j} \log(\vet{C}_a(\vet{M}_a(\vet{x}_i)))\\
		           & + (1 - a_{i,j}) \log (1-\vet{C}_a(\vet{M}_a(\vet{x}_i))) \Big),
	\end{split}
\end{equation}
in which $\vet{C}_a$ is the attribute classifier, as in Fig. \ref{fig: attribute_recognition} and $\vet{M}_a$ denotes the feature mapping in \rul{$\mathcal{D}_a$}. The attribute recognition network is trained by minimizing the loss $L_{attr}$.

\vspace{2mm}
\noindent
\textbf{Unsupervised Domain Adaptation:} Given the learned mapping $\vet{M}_a$ in the attribute recognition domain $\mathcal{D}_a$, the objective of this step is to learn a domain invariant mapping $\vet{M}$ for attribute-related feature extraction in the ReID domain $\mathcal{D}_p$. Since the attribute label is unavailable in $\mathcal{D}_p$, an unsupervised domain adaptation method is used to adapt the network trained in $\mathcal{D}_a$ to the new domain $\mathcal{D}_p$. We modify ADDA \cite{tzeng2017adda} to make it more suitable for network adaptation.

In ADDA, the classification network is \rul{first} trained in source domain with label supervision. To adapt the trained source network to the target domain, an adversarial adaptation strategy is used to learn the target mapping. A good adapted target mapping means the discriminator network $\vet{D}$ can not reliably predict the domain label of the feature vector from source or target mapping. As in generative adversarial network (GAN), the adapted target encoder is trained by playing the minimax game:
\begin{equation}
\label{eq:adda_minimax}
\begin{split}
	\min_{\vet{M}_t} \max_{\vet{D}} L_{adv} =&- \mathbb{E}_{\vet{x}_i \sim \vet{X}^{s}}[\log \vet{D}(\vet{M}_s(\vet{x}_i))] \\
                                 &- \mathbb{E}_{\vet{x}_i \sim \vet{X}^{t}} [\log (1-\vet{D}(\vet{M}_t(\vet{x}_i))],
\end{split}
\end{equation}
in which $\vet{M}_t$ is the feature mapping in the target domain (usually initialized with the weights from $\vet{M}_s$). Similarly, $\vet{M}_s$ is the pretrained source feature mapping and it is fixed during adaptation.

Since in ADDA only samples from the target domain are fed into the target network, the latter is not domain invariant and it can be only used in the target domain. To learn a domain invariant mapping, we use samples from both the target (ReID) and source (attribute recognition) domain as in \cite{volpi2017adversarial}. There are at least two advantages to do this. Firstly, the target mapping is invariant to both the attribute recognition and ReID domains and thus the extracted features are more robust to the domain shift compared to the original ADDA. The second one is that more samples are used to train the target encoder. In our experiments, we found that this makes the modified adversarial adaptation easier to converge than the original one. The least square GANs, which uses the least square loss function for the discriminator, has been reported as a stable variation of GANs \cite{volpi2017adversarial, mao2017lsgan}. In the modified adversarial adaptation, the least square loss function is also used in the adversarial loss $L_{adv}$ as:
\begin{equation}
  \begin{split}
	\min_{\vet{M}} \max_{\vet{D}} L_{adv} =& \mathbb{E}_{\vet{x}_i \sim \vet{X}^{a}\,\cup \,\vet{X}^p} \Arrowvert \vet{D}(\vet{M}(\vet{x}_i)) -1 \Arrowvert ^2 \\
	&+ \mathbb{E}_{\vet{x}_i \sim \vet{X}^{a}} \Arrowvert \vet{D}(\vet{M}_a(\vet{x}_i)) \Arrowvert^2,
  \end{split}
\end{equation}
in which $\vet{X}^p = [\vet{x}_1^p, \ldots, \vet{x}_{n_p}^p]$ and $\vet{M}$ is initialized with the weights from $\vet{M}_a$, that is fixed during training. 

With the domain invariant $\vet{M}$, we can now extract attribute-discriminative features in $\mathcal{D}_p$. In our experiments, we also observed that the attribute recognition performance of the original ADDA decreases during the domain adaptation procedure. In order to cope with this drop in performance, an additional attribute classifier is added for source samples, as shown in Fig. \ref{fig: adversarial adaptation}. Thus, the final optimization problem is:
\begin{align}
\label{eq: proposed_adaptation_loss}
	\min_{\{\vet{M}, \vet{C}\}} \max_{\vet{D}} L_{adv} + \alpha L_{attr},
\end{align}
where $\alpha$ is a hyper-parameter. $L_{attr}$ is the same as in Eq. \eqref{eq:ar_loss} with $\vet{C}, \vet{M}$ in place of $\vet{C}_a, \vet{M}_a$. The former weights are initialized with the latter, as often done in domain adaptation. 

\vspace{2mm}
\noindent
\textbf{Person ReID:} \rul{After domain adaptation, the target CNN (or feature mapping $\vet{M}$)} \rul{is used for extracting attribute-related features of person images in the ReID domain $\mathcal{D}_p$. The similarities between these extracted features are calculated simply using the Euclidean distance as in deep ReID works} \cite{zheng2016reid_survey, zheng2016mars, wang2018transferable}. \rul{Finally, matching or ranking person images can be performed based on their similarities.}

\section{Experiments}
\label{sec:exps}

In this section, experimental results are presented to demonstrate the effectiveness of the proposed ReID feature learning framework. To measure the performances of the proposed method, the Cumulative Matching Characteristic (CMC) and mean average precision (mAP) are used.

\vspace{2mm}
\noindent
\textbf{Datasets:} Three datasets, \ie RAP, Market-1501 and Duke-MTMC-reID, are included to evaluate the proposed ReID framework \cite{zheng2016reid_survey, RAP}. RAP is an attribute recognition dataset that includes $41, 585$ images in total. For each image $91$ pedestrian attributes have been annotated. In our experiments we removed some extremely unbalanced attributes, selecting only 70 attributes. Market-1501 and DukeMTMC-ReID are two popular large-scale ReID datasets. Thanks \rul{to} the work \cite{lin2017attribute_recognition}, they are also labeled with pedestrian attributes, \rul{specifically} $27$ attributes in Market-1501 and $22$ in DukeMTMC-reID.

\vspace{2mm}
\noindent
\textbf{Implementation details:} In all experiments, the input images are resized to $(224, 224, 3)$. The hyper-parameter $\alpha$ in Eq. \eqref{eq: proposed_adaptation_loss} is fixed to 0.1. MobileNet is selected as the backbone network and it is pretrained on ImageNet. Adam optimizer with a learning rate of 0.0001 is used in all experiments. Two fully connected (FC) layers are included in the attribute classifier (1024 $\rightarrow$ 512 $\rightarrow$ $m$). The discriminator is also composed of two FC layers (1024 $\rightarrow$ 384 $\rightarrow$ 1 or 1024 $\rightarrow$ 128 $\rightarrow$ 1 for the adaptation with or without classifier).


\begin{table}[t]
\caption{Comparing experimental results before and after adaptation. The network is \rul{first} trained on the attribute recognition dataset RAP.}
\center
\begin{tabular}{ccccc}
\hline
                       & \multicolumn{2}{||c|}{Market-1501}                        & \multicolumn{2}{||c}{DukeMTMC-reID}\\ \hline\hline
Metric                 & \multicolumn{1}{||c|}{Rank1} & \multicolumn{1}{c}{mAP}    & \multicolumn{1}{||c|}{Rank1} & \multicolumn{1}{c}{mAP} \\ 
\hline
\multicolumn{1}{c}{w/o adaptation} & \multicolumn{1}{||c|}{28.2\%} & \multicolumn{1}{c}{8.7\%} & \multicolumn{1}{||c|}{15.6\%} & \multicolumn{1}{c}{4.9\%} \\ 
\hline
\multicolumn{1}{c}{w adaptation}  & \multicolumn{1}{||c|}{\textbf{32.1\%}} & \multicolumn{1}{c}{\textbf{10.6\%}} & \multicolumn{1}{||c|}{\textbf{18.7\%}} & \multicolumn{1}{c}{\textbf{6.5\%}} \\ 
\hline
\end{tabular}
\label{tb:exp_before_after_adaptation}
\end{table}


\begin{table}[t]
\caption{Performance comparisons with existing attribute based unsupervised domain-adaptive person ReID methods.}
\center
\begin{tabular}{ccccc}
\hline
Source $\rightarrow$  Target & \multicolumn{4}{||c}{Market-1501 $\rightarrow$  DukeMTMC-reID} \\ \hline
Metric                 & \multicolumn{1}{||c|}{Rank1} & \multicolumn{1}{c|}{Rank5}  & \multicolumn{1}{c||}{Rank10}  & \multicolumn{1}{c}{mAP} \\ 
\hline
\multicolumn{1}{c}{TJ-AIDL} 
& \multicolumn{1}{||c|}{24.3\%} 
& \multicolumn{1}{c|}{38.3\%} & \multicolumn{1}{c||}{45.7\%} &  \multicolumn{1}{c}{10.0\%} \\ 
\hline
\multicolumn{1}{c}{MMFA}  & \multicolumn{1}{||c|}{15.8\%} & \multicolumn{1}{c|}{26.0\%} & \multicolumn{1}{c||}{48.2\%} &  \multicolumn{1}{c}{5.7\%} \\ 
\hline
\multicolumn{1}{c}{\textbf{Ours}}  & \multicolumn{1}{||c|}{\textbf{28.6\%}} & \multicolumn{1}{c|}{\textbf{44.2\%}} & \multicolumn{1}{c||}{\textbf{51.7\%}} &  \multicolumn{1}{c}{\textbf{13.1\%}} \\ 
\hline
\hline 
Source $\rightarrow$  Target & \multicolumn{4}{||c}{DukeMTMC-reID $\rightarrow$  Market-1501} \\ 
\hline
Metric                 & \multicolumn{1}{||c|}{Rank1} & \multicolumn{1}{c|}{Rank5}  & \multicolumn{1}{c||}{Rank10}  & \multicolumn{1}{c}{mAP} \\ 
\hline
\multicolumn{1}{c}{TJ-AIDL} & \multicolumn{1}{||c|}{38.0\%} & \multicolumn{1}{c|}{59.2\%} & \multicolumn{1}{c||}{67.6\%} &  \multicolumn{1}{c}{13.6\%} \\ 
\hline
\multicolumn{1}{c}{MMFA}  & \multicolumn{1}{||c|}{35.5\%} & \multicolumn{1}{c|}{55.3\%} & \multicolumn{1}{c||}{64.0\%} &  \multicolumn{1}{c}{12.7\%} \\ 
\hline
\multicolumn{1}{c}{\textbf{Ours}}  & \multicolumn{1}{||c|}{\textbf{43.0\%}} & \multicolumn{1}{c|}{\textbf{63.3\%}} & \multicolumn{1}{c||}{\textbf{70.6\%}} &  \multicolumn{1}{c}{\textbf{17.1\%} }\\ 
\hline
\hline

\end{tabular}
\label{tb:comparison_with_soa}
\end{table}

\vspace{2mm}
\noindent
\textbf{Experimental results and discussions:} To evaluate the performance of the adaptation procedure in the proposed unsupervised adaptive ReID framework, Table \ref{tb:exp_before_after_adaptation} shows the experimental results before and after domain adaptation. The network is \rul{first} trained on \rul{RAP}. For the experiments \rul{without} adaptation, the trained attribute recognition network is directly used on ReID dataset, \ie Market-1501 or DukeMTMC-reID, to extract ReID feature representations. From the table, it can be found that after adaptation, the ReID performance of the network has been improved. There are $3.9\%$ Rank1 improvements on Market-1501 and $3.1\%$ on DukeMTMC-reID. \rul{This validates the effectiveness of the proposed unsupervised adversarial adaptation for ReID}. An interesting observation from \rul{Table} \ref{tb:exp_before_after_adaptation} is that both \rul{Rank1} and mAP metrics of Market-1501 are much higher than DukeMTMC-reID. This is mainly \rul{because} there are large domain differences between \rul{RAP and DukeMTMC-reID while it is small between RAP and Market-1501}.

\begin{figure}[t]
\centering
\includegraphics[width=0.35\textwidth]{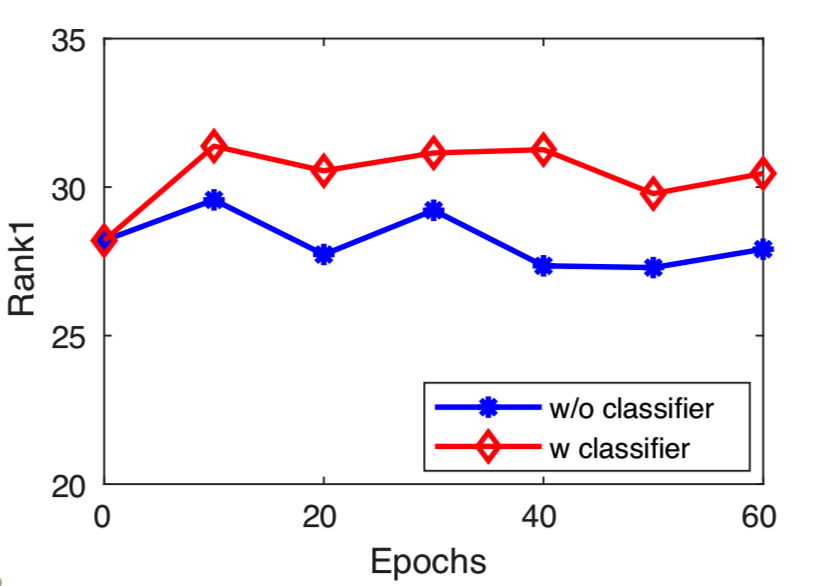}\\
\caption{Adaptation performance comparisons between the unsupervised domain adaptive ReID frameworks with and without the additional classifier. 0 epoch means there is no adaptation in the proposed ReID framework.}
\label{fig: wo_w_classifier}
\end{figure}

\rul{Fig.} \ref{fig: wo_w_classifier} shows the experimental results of the proposal ReID framework with and without the additional classifier. The experiments are performed on RAP and Market-1501 datasets. The network is \rul{first} trained on RAP dataset and then adapted to Market-1501 dataset. For the ReID framework without classifier, \rul{the classifier has been deleted in the adaptation step.} In the experiments, we found the attribute recognition performance degrades during \rul{the adaptation of the network}. As observed in Fig. \ref{fig: wo_w_classifier}, this results in the decrease of the ReID performance during domain adaptation \rul{because of the strong relation between attribute recognition and ReID}. With the additional classifier, the ReID performance, as the red curve in the figure, are better and more stable.

In order to compare with state-of-arts, the experiments using Market-1501 and DukeMTMC-reID datasets are also performed. As in TJ-AIDL \cite{wang2018transferable} and MMFA \cite{lin2018multi}, one of the datasets is used as the source domain and the other one is for target domain. For fair comparison, only attribute based experimental results of TJ-AIDL and MMFA are selected. \rul{From Table} \ref{tb:comparison_with_soa}, it shows our proposed method outperforms both \rul{TJ-AIDL and MMFA}. For example, in the case of Market-1501 $\rightarrow$  DukeMTMC-reID, there are $4.3\%$ improvements on Rank1 compared with TJ-AIDL, and $12.8\%$ improvements compared with MMFA. In addition, \rul{Table} \ref{tb:comparison_with_soa} also shows that the ReID performance in the case of DukeMTMC-reID $\rightarrow$  Market-1501 is better than Market-1501 $\rightarrow$  DukeMTMC-reID (from $28.6\%$ to $43.0\%$ for Rank1 \rul{of} our proposed method). \rul{This is due to the fact that compared with Market-1501, DukeMTMC-reID contains more variations in samples, for example more changes in image resolution and background clutter} \cite{DukeMTMC-reID, wang2018transferable}, \rul{and thus the network trained on it has better generalization ability.}


\section{Conclusions}
\label{sec:conclu}

In this work, we presented an unsupervised domain adaptive ReID framework for extracting attribute-related features. Considering that \rul{most of ReID datasets are not labeled with attributes, an unsupervised domain adaptation method has been proposed for learning domain invariant attribute-related features, which can be used for ReID in the target domain.} Based on the observation that the attribute recognition performance degrades during domain adaptation, an additional classifier has been added. Experimental results using three large-scale datasets proved the effectiveness of the proposed approach.

\newpage
\bibliographystyle{IEEEbib}
\bibliography{main.bib}


\end{document}